# A Preliminary Study for Building an Arabic Corpus of Pair Questions-Texts from the Web: AQA-Webcorp

http://dx.doi.org/10.3991/ijes.v4i2.5345

Wided Bakari[1], Patrice Bellot[2], Mahmoud Neji[1]
[1] Faculty of Economics and Management, Sfax, Tunisia
[2] Aix-Marseille University, Marseille, France

*Abstract*—With the development of electronic media and the heterogeneity of Arabic data on the Web, the idea of building a clean corpus for certain applications of natural language processing, including machine translation, information retrieval, question answer, become more and more pressing. In this manuscript, we seek to create and develop our own corpus of pair's questions-texts. This constitution then will provide a better base for our experimentation step. Thus, we try to model this constitution by a method for Arabic insofar as it recovers texts from the web that could prove to be answers to our factual questions. To do this, we had to develop a java script that can extract from a given query a list of html pages. Then clean these pages to the extent of having a data base of texts and a corpus of pair's question-texts. In addition, we give preliminary results of our proposal method. Some investigations for the construction of Arabic corpus are also presented in this document.

## I. INTRODUCTION

The lack and / or the absence of corpus in Arabic has been a problem for the implementation of natural language processing. This also has a special interest in the track of the question answering.

The corpus (singular form of corpora[1]) construction is a task that both essential and delicate. It is complex because it depends in large part a significant number of resources to be exploited. In addition, the corpus construction is generally used for many NLP applications, including machine translation, information retrieval, question-answering, etc. Several attempts have succeeded of building their corpus. According to [Sinclair, 2005], a corpus is a collection of pieces of texts in electronic forms, selected according to external criteria for end to represent, if possible, a language as a data source for linguistic research. Indeed, a definition that is both specific and generic of a corpus according [Rastier, 2005], is the result of choices that brings the linguists. A corpus is not a simple object; it should not be a mere collection of phrases or a "bag of words". This is in fact a text assembly that can cover many types of text.

With the internet development and its services, the web has become a great source of documents in different languages and different areas. This source is combined with storage media that allow the rapid construction of a corpus [Meftouh et al., 2007]. In addition, using the Web as a base for the establishment of textual data is a very recent task. The recent years have taken off work attempting to exploit this type of data. From the perspective of automated translation in [Resnik et al., 1998], the others study the possibility of using the websites which offering information in multiple languages to build a bilingual parallel corpus.

Arabic is also an international language, rivaling English in number of native speakers. However, little attentions have been devoted to Arabic. Although there have been a number of investigations and efforts invested the Arabic corpus construction, especially in Europe. Progress in this area is still limited. In our research, we completed building our corpus of texts by querying the search engine Google. We are concerned our aim, a kind of, giving a question, analyzing texts at the end to answer this question.

This paper is organized into six sections as follows: it begins with an introduction, followed by use of the web as a corpus source. Section 3 outlines the earlier work in Arabic; Section 4 shows our proposed approach to build a corpus of pairs of questions and texts; Section 5 describes an experimental study of our approach; a conclusion and future work will conclude this article.

## II. USING THE WEB AS A SOURCE OF CORPUS

Although the web is full of documents, finding information for building corpus in Arabic is still is a tedious task. Nowadays, the web has a very important role in the search for information. It is an immense source, free and available. The web is with us, with a simple click of the mouse, a colossal quantity of texts was recovered freely [Gatto, 2011]. It contains billions of text words that can be used for any kind of linguistic research [Kilgarriff & Grefenstette, 2001]. It is considered the greatest knowledge resource. Indeed, the current search engines cannot find extracts containing the effective answers to a user question. Which sometimes makes it difficult to get accurate information; sometimes, they can't even extract the correct answers. The web is also an infinite source of resources (textual, graphical and sound). These resources allow the rapid constitution of the corpus. But their constitution is not easy so that it raises a number of questions [Isaac, 2001].

---

[1] McEnery (2003) pointed out that corpuses is perfectly acceptable as a plural form of corpus.

The construction of a corpus of texts from the web was not a simple task. Such constitution has contributed to developing and improving several linguistic tools such as question-answering systems, information extraction systems, machine translation systems, etc.

In fact, one of the most interesting Web intrinsic characteristics is its multilingual. As for the current distribution of languages used on the Web, recent estimates of the top ten languages (30 June 2015[2]) report that English and Chinese are the most used languages, followed by Spanish. The Arabic is a fourth on Internet users by language, followed by other major languages such as Portuguese, Japanese, Russian, German, Malay, French, and German.

Our intent is threefold. First, the empirical evaluation must be based on a relevant corpus. Secondly, we seek to analyze our built corpus. Finally, the main purpose of our research concerns the search for an accurate answer to a question in natural language. That is why we considered the Web as a source of potential to build our corpus.

### III. LITERATURE REVIEW: ARABIC CORPUS CONSTRUCTION FROM THE WEB

The corpus is a resource that could be very important and useful in advancing the various language applications such as information retrieval, speech recognition, machine translation, question-answering, etc. This resource has gained much attention in NLP. The task of building a corpus of textual resources from the web is somewhat recent. In Arabic, the attempts to exploit this type of data are limited. Although there has been some effort in Europe, which led to the successful production of some Arabic corpus; Progress in this field is still limited. According to [Atwell et al 2004], the progress has been hampered by the lack of effective corpus analysis tools, such as taggers, stemmers, readable dictionaries to the machine, the corpus viewing tools, etc., that are required for build and enrich a corpus as a research tool.

Many question-answering systems rely on document corpus construction in English and other languages; there are some publicly accessible corpuses, especially in Arabic. This language has not received the attention that it deserves in this area. In this regard, many researchers have emphasized the importance of a corpus and the need to work on their construction. For his part, [Mansour, 2013], in his work, showed that the contribution of a corpus in a linguistic research is a huge of many ways. In fact, as such a corpus provides an empirical data that enables to form the objective linguistic declarations rather than subjective. Thus, the corpus helps the researcher to avoid linguistic generalizations based on his internalized cognitive perception of language. With a corpus, the qualitative and quantitative linguistic research can be done in seconds; this saves the time and the effort. Finally, the empirical data analysis can help the researchers not only to precede the effective new linguistic research, but also to test the existing theories.

In this section, we present the most significant studies on English corpus construction and previous attempts to build a corpus in Arabic. In addition, we will also cover the studies that claim the construction of corpus for question-answering. There are now numerous studies that use the Web as a source of linguistic data. Here, we review a few studies for the Arabic language, as our case, that use the search engines queries to build a corpus.

Among the most recognized Arabic corpus construction projects, we cite, for example, the work of Resnik which studies the possibility of using the websites offers the information's in multiple languages to build the bilingual parallel corpora [Resnik, 1998].

Ghani and his associates [Ghani et al., 2001] performs a study of building a corpus of minority languages from the web by automatically querying the search engines.

In order to study the behavior of predicate nouns that highlight the location and movement, the approach proposed by Isaac and colleagues [Isaac et al., 2001] developed software for the creation of a corpus of sentences to measure if the introduction of prepositions in queries, in information retrieval, can improve the accuracy.

Even more, the work of [Baroni & Bernardini, 2004] introduced the "BOOTCAT Toolkit". A set of tools that allow an iterative construction corpus by automated querying the Google and terminology extraction. Although it is devoted to the development of specialized corpora, this tool was used by [Ueyama & Baroni, 2005] and [Sharoff, 2006] to the generalized corpus constitution.

Similarly, the work of [Meftouh et al., 2007] describes a tool of building a corpus for the Arabic. This corpus automatically collected a list of sites dedicated to the Arabic language. Then the content of these sites is extracted and normalized. Indeed, their corpus is particularly dedicated for calculating the statistical language models.

In another approach of Elghamry in which he proposed an experiment on the acquisition of a corpus from the web of the lexicon hypernymy-hyponymy to partial hierarchical structure for Arabic, using the pattern lexico-syntactic "بعض x مثل y1 ... yn" (certain x as y1, ... yn) of [Elghamry, 2008].

From the perspective of automatic summarization, [Maâloul et al., 2010] studied the possibility of extracting Arabic texts of the website "Alhayat" by selecting newspaper articles of HTML type with UTF-8 encoding, to locate the rhetoric relations between the minimum units of the text using rhetorical rules. In addition, [Ghoul, 2014] provides a grammatically annotated corpus for Arabic textual data from the Web, called Web Arabic corpus. The authors note that they apply the "Tree tagger" to annotate their corpus based on a set of labels. This corpus consists of 207 356 sentences and 7 653 881 words distributed on four areas: culture, economy, religion and sports.

Finally, in [Arts et al., 2014] the authors present arTenTen, an Arabic explored corpus from the web. This one is a member of the family of Tenten corpus [Jakubíček et al., 2013]. arTenTen consists of 5.8 billion words. Using the JusText and Onion tools, this corpus has been thoroughly cleaned, including the removal of duplicate [Pomikalek, 2011]. The authors use the version of the tool MADA 3.2 for marking task, lemmatization and part-of-speech tagging of arTenTen ([Habash & Rambow, 2005] [Habash et al., 2009]). This corpus is compared to two other Arabic corpus Gigaword [Graff, 2003] and an explored corpus of the web [Sharoff, 2006].

Documents on the web have also operated by other approaches and other researchers ([Volk, 2001] [Volk, 2002], [Keller & Lapata, 2003], [Villasenor-pineda et al., 2003]) to address the problem of the lack of data in statis-tical modeling language [Kilgarrif & Grefenstette, 2003]. In [Elghamry et al., 2007], these data were used to resolve anaphora in Arabic free texts. Here, the authors construct a dynamic statistical algorithm. This one uses the fewest possible of functionality and less human intervention to overcome the problem lack sufficient resources in NLP. However, [Sellami et al., 2013] are working on, the online encyclopedia, Wikipedia to retrieve the comparable articles.

---

[2] www.internetworldstats.com

In a merge of search engine and language processing technologies, the web has also been used by groups in Sheffield and Microsoft among others as source of answers for question-answering applications ([Mark et al., 2002], [Susan et al., 2002]). AnswerBus [Zhiping, 2002] allows answering the questions in English, German, French, Spanish, Italian and Portuguese.

Although the text corpus building efforts are focused on English, Arabic corpus can also be acquired from the Web which is considered as a large data source. These attempts might be in all of the NLP applications. However, we also note significant efforts mainly for the question-answering. In this regard, the major of our knowledge, the number of corpus dedicated to Arabic question-answering is somewhat limited. Among the studies that have dedicated to this field, we cite [Trigui et al., 2010] who built a corpus of definition questions dealing the definitions of organizations. They use a series of 50 organization definition questions. They experienced their system using 2000 extracts returned by the Google search engine and Arabic version of Wikipedia.

We conclude that the natures of web texts are liable to be show up in many applications of automatic processing Arabic language. There are, at present, a number of construction studies of texts corpus in various applications, including the named entity recognition, plagiarism detection, parallel corpus, anaphora resolution, etc. The efforts to build corpus for each application are significant. They could be in all NLP applications.

## IV. THE CHALLENGES OF THE ARABIC LANGUAGE

Although, Arabic is within the top ten languages in the internet, it lacks many tools and resources. The Arabic does not have capital letters compared the most Latin languages. This issue makes so difficult the natural language processing, such as, named entity recognition. Unfortunately there is very little attention given to Arabic corpora, lexicons, and machine-readable dictionaries [Hammo et al., 2002]. In their work [Bekhti & Al-harbi, 2011], the authors suggest that the developed Arabic question-answering systems are still few compared to those developed for English or French, for instance. This is mainly due to two reasons: lack of accessibility to linguistic resources and tools, such as corpora and basic Arabic NLP tools, and the very complex nature of the language itself (for instance, Arabic is inflectional and non concatenative and there is no capitalization as in the case of English). On their part, [Abdelnasser et al., 2014] illustrate some difficulties of Arabic. This language is highly inflectional and derivational, which makes its morphological analysis a complex task. Derivational: where all the Arabic words have a three or four characters root verbs. Inflectional: where each word consists of a root and zero or more affixes (prefix, infix, suffix). Arabic is characterized by diacritical marks (short vowels), the same word with different diacritics can express different meanings. Diacritics are usually omitted which causes ambiguity. Absence of capital letters in Arabic is an obstacle against accurate named entities recognition. And then, In their survey [Ezzeldin & Shaheen, 2012], the authors emphasize that as any other language, Arabic natural language processing needs language resources, such as lexicons, corpora, treebanks, and ontologies are essential for syntactic and semantic tasks either to be used with machine learning or for lookup and validation of processed words.

## V. PRESENTATION OF AQA-WEBCORP CORPUS: ARABIC QUESTION ANSWERING WEB CORPUS

In the rest of this article, we show in detail our suggestion to build our corpus of pair's questions and texts from the web as well as our preliminary empirical study. In our case, the size of the corpus obtained depends mostly on the number of questions asked and the number of documents selected for each question.

In the context of their construction of a corpus of texts from the web, researchers in [Issac et al., 2001] point out that there are two ways to retrieve information from the web for building a corpus. The first one is to group the data located on known sites [Resnik, 1998]. Indeed, this way runs a vacuum cleaner web. This ensures the recovery of the pages from a given address. However, the second method investigates a search engine to select addresses from one or more queries (whose the complexity depends on the engine). Thereafter, recover manually or automatically the corresponding pages from these addresses.

In our work we follow the second method. From a list of questions posed in natural language, we ensure the recovery of the list of corresponding URLs. Then, from these URLs, we propose to recover the related web pages. Eventually, we propose to clean these pages so as to produce the lists of texts that will be the foundation that built our corpus. After building our corpus of pair's questions and texts, we do not keep it to that state. In this respect, a stage of analysis and processing will be looked later to achieve our main objective which is the extraction of an adequate and accurate answer to each question.

### A. Collection of the questions

It is comes to collect a set of questions in natural language. These questions can be asked in different fields, including sport, history & Islam, discoveries & culture, world news, health & medicine.

Currently, our corpus consists of 115 pairs of questions and texts. Indeed, the collection of these Questions is carried out from multiple sources namely, discussion forums, frequently asked questions (FAQ), some questions translated from the two evaluation campaigns TREC and CLEF (Figure 1).

The data collected from the web of the questions and the texts will help us to build an extensible corpus for the Arabic question-answering. The size of our corpus is in the order of 115 factual questions: 10 questions translated from TREC, 10 questions translated from CLEF and 95 questions gathered from the forums and FAQs. To build our corpus, we used the Arabic texts available on the Internet that is collected being based on the questions posed at the outset.

From a perspective of analysis and post processing carried out to our corpus taking into account the form and the content of the corresponding questions, we collected factual questions having the type (What, Where, When, Who, How) (ما , أين, متى, من كم) (Figure 2).

### B. The steps of construction

Most studies in Arabic corpus construction are designed for areas other than the question-answering (see figure). According to a research done at Google3, we find that the number of attempts devoted to this area is so limited. That's why we decided to build our own corpus. To address this goal we need as an intermediate step interrogating a search engine. The construction of the corpus for the question-answering gets better. We hope that it will continue to improve in the coming years and will complete one day to produce corpus in this field of research that will be used by researchers in their empirical studies.

Our approach is described in [Bakari et al., 2014] and [Bakari et al., 2015]; The first stage of this approach is the question analysis. Indeed, in this module, we collect and analyse our questions to generate some features for each one. Essentially, these features are the list of question keywords, the focus of the question, and the expected answer type. With a real interrogation of Google, these characteristics may recover for each given question the extracts that address answers for this question. Then, this module infers the question in his declarative form in order to generate, as much as possible, logical representation for each question. Furthermore, the different modules of the extracting the accurate presented in our approach are strongly linked to the question analysis module. In one hand, the keywords and the focus are used to interrogate Google and recuperate relevant passages. In other hand, the declarative form is designed to infer the logic representation. And then, the expected answer type is carried out to select the accurate answer.

To implement our corpus for Arabic, we propose a simple and robust method implemented in Java. The principle of this method is based on four stages, relatively dependent. The constitution of our corpus of pairs of Arabic question and texts is actually done by developing all of these four steps. We describe in the following each of these steps:

- Documents research
- Web pages recovery
- Texts preparation
- Texts classification

This methodological framework is to look for web addresses corresponding to each question. Indeed, we have segmented questions in a list of keywords. Then our tool seeks list of URLs addresses which match those keys words. Then, for each given address we propose to recover the webpage that suits him. In this respect, our corpus construction tool is an interface between the user request and Google. Specifically, it is a way to query the Google database to retrieve a list of documents. Finally, we performed a transformation of each retaining web page from (".html" ➔ ".txt"). Finally, we look for if the answer is found in the correspondent text. The text is considered valid to build our body if it contains this answer. Otherwise, we go to the following URL.

As illustrated in the (figure 4), we introduce in this section a simple method of construction of our corpus pro-moting an effective interrogation of the web. This method is generally composed of three modules. Given a module for generating list of URLs address is implemented, this module supports for any questions posed in natural language a list of corresponding URLs. Next, a processing module behaves like a corresponding web page generator. A third module provides a sort of filtering of these pages. The result of this module could be a set of texts that can be added to the questions to build our corpus from the web.

---

3 http://www.qatar.cmu.edu/~wajdiz/corpora.html

## VI. EMPIRICAL ASSESSMENT

The first stage consists from a question posed in natural language, to attribute the list of URLs corresponding addresses to it. Indeed, the documents search is done being based on the words of the collected questions. A better to ensure this step we developed a java script for interrogating Google to obtain these results. The result of this step is a set of URLs addresses. For each question a list of the URLs will be affected.

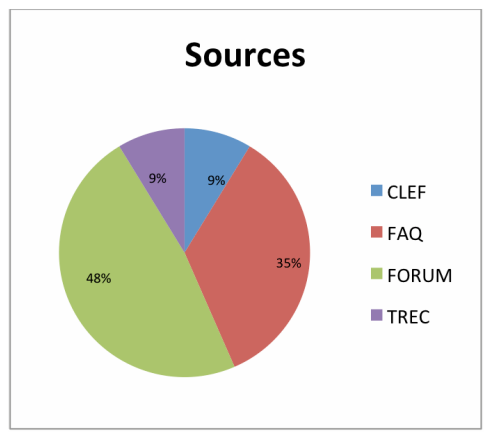

Figure 1.  Source of the questions used for our corpus

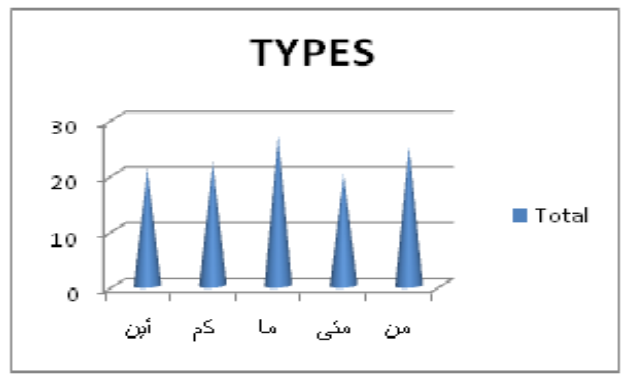

Figure 2.  Some examples of questions used in our corpus

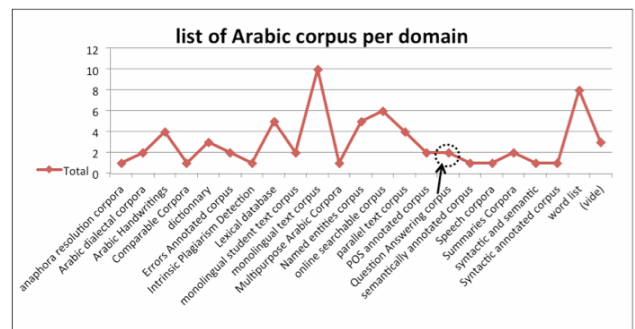

Figure 3.  List of Arabic corpus per domain

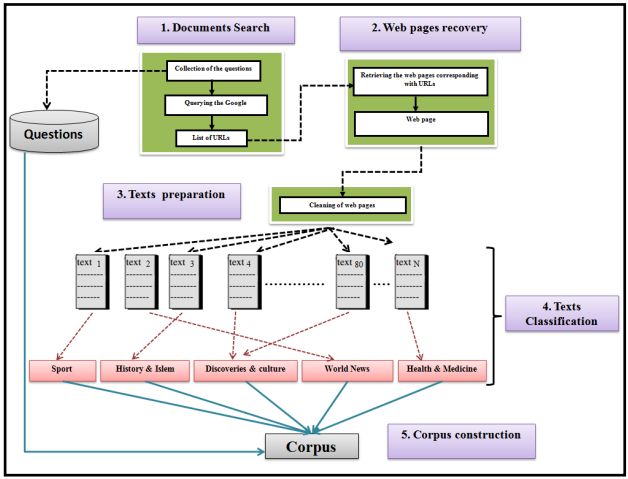

Figure 4. Construction process of our corpus AQA-WebCorp

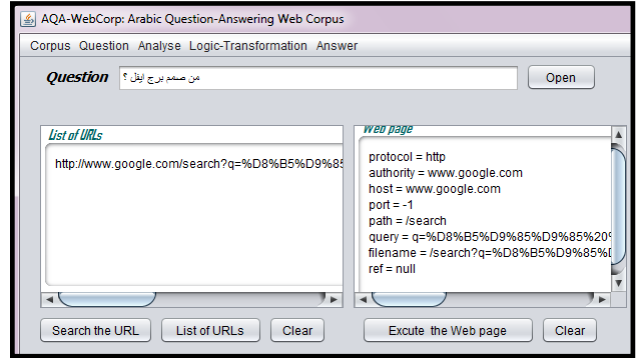

Figure 5. List of URLs generated for the question " من صمم برج ايفل ؟ ".

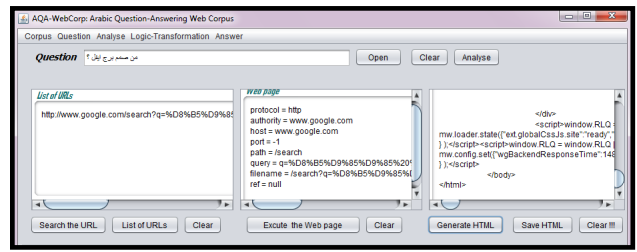

Figure 6. An URL of a web page retaining from the question " من صمم برج ايفل ؟ ".

In our case, to look for an answer to a question in Arabic, we propose to use a search engine (e.g. Google) to retrieve the documents related to each question. Then add post linguistic treatments to those documents which are actually constitute our corpus to have an accurate and appropriate answer. In this respect, querying a search engine accelerates the recovery of documents online but requires an offline processing of these documents. We think that this is a better solution, but much more complex is the implementation of a Linguistic search engine for a particular purpose.

At this stage, the module of documents search is implemented. First, when a question is asked, our tool submits it to the search engine (Google) to identify the list of URLs based on a list of words constitute this question.

Consider the following example: our tool can then from the question: « من صمم برج ايفل ؟ » Generate a list of equivalent URLs addresses. The default Web access means is through a search engine such as Google. In addition, by clicking the "search URLS" button, a list of addresses will be automatically displayed. While for each URL, this prototype can retrieve the necessary information's (host, query, protocol, etc.).

This module, developed in Java, describes our desired process of the passage retrieval from the Web. As a starter, it takes 115 factual questions of five categories (Who, What, When, Where, How). For each question, the elements that it characterizes are recovered: Focus of the question, the expected answer type, the list of keywords, and then its declarative form. Of this declarative form, a logical representation that it suits is established. With a real interrogation of Google, text passages that contain the answers to these questions are chosen. For each question, we assume to be near 9 passages are retrieved. Those passages, in their totality, constitute a text. Besides, the set of texts recovered with the collected questions, build our AQAWebCop corpus. Currently, we are interested in analyzing these texts. To do this, we propose, first, to segment each text into a list of sentences. On the other hand, we aim to transform the sentences into logical forms. In our study of looking for a specific answer, the logic is presented in the question analysis as in the text analysis.

Once the list of URLs is generated, our tool must determine for each address the corresponding web page. This is to look for the corresponding HTML page for each given URL. The following figure illustrates this case. From the address retained in the first step, a set of web pages is recovered. Each of these pages is exported in the format ".html".

The figure below is an actual running of our prototype, using the previous steps of figure 6. The "generate HTML" button retrieves a page from its URL (figure 7).

The last step is to transform every web page obtained in the previous step into a ".txt" format. The texts being in ".html" format, and given that the intended application is the statistical language modeling, it seems justified to put them in the ".txt" format. For this, we remove all the HTML tags for each retrieved pages. As we have said before, our method seeks answers to each question in each generated text. It is possible either is to keep the text for own corpus construction work, or to disregard it.

When the text is generated it shall be ensured its suitability to answer the question put in advance, before saving it. To select this text, we proposed to select from the retained text list one which is to compile the most of information's related to the question. When this text is found, it is saved at the end of to be able to build our own corpus.

We are currently developing a corpus dedicated to the Arabic question-answering. The size of the corpus is in the order of 115 pairs of questions and texts. This is collected using the web as a source of data. The data collected, of these questions and texts from the web, will help us to build an extensible corpus for the Arabic question-answering. The pairs of texts-questions distributed on five areas "صحة و ; رياضة; ثقافةو إكتشافات; التاريخ والإسلام; أخبار العالم" "طب" as shown in Figure 10.

Through the Google search engine, we develop a prototype that builds our corpus to Arabic question-answering. This corpus is in the order of 115 pairs of questions and texts. We are experimented our tool using 2875 URLs re-turned by the search engine Google. This is due to the fact that the number of URLs to each particular question is approximately between 23 and 26 URLs. Currently our corpus contains 115 pairs of questions and texts.

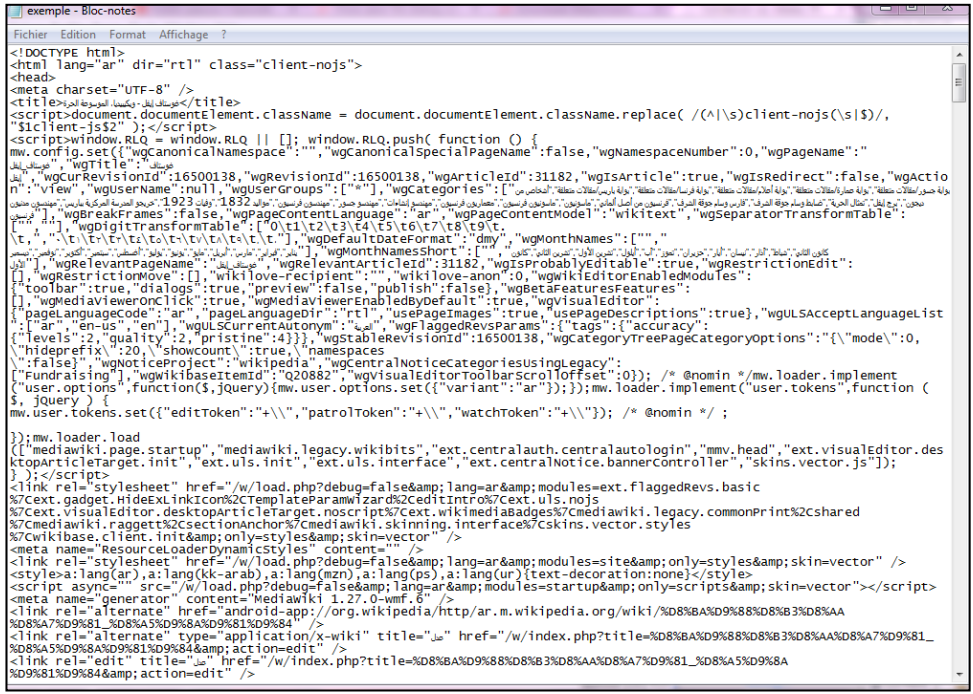

Figure 7.   HTML page generated

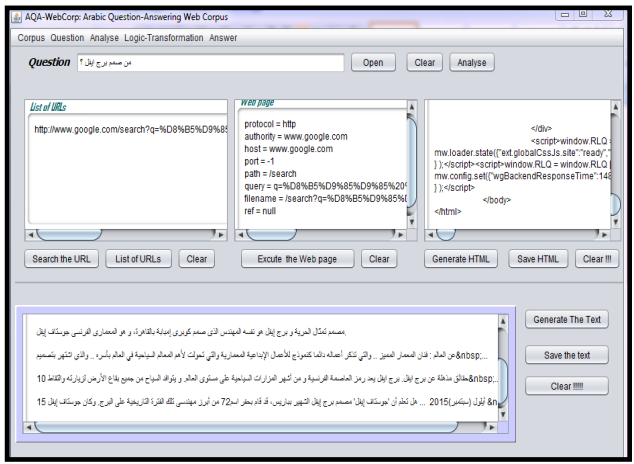

Figure 8.   Example of a text containing an answer to the following question:: " من صمم برج ايفل ؟ ".

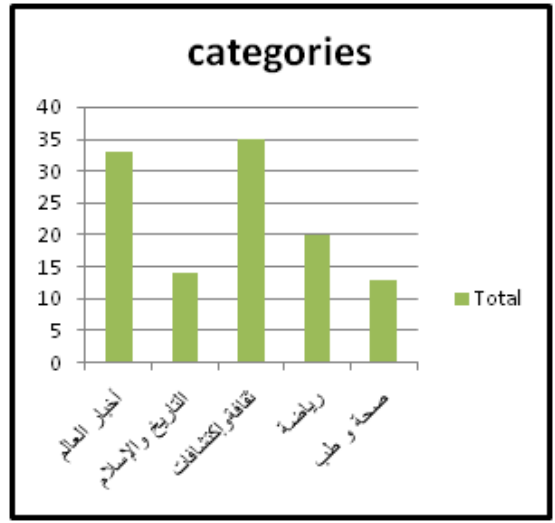

Figure 10. Statistics of pairs questions-texts used in different sources

The evaluation of our corpus strongly depends on the number of URLs correctly reported by Google. In this regard, we completed an intrinsic evaluation on a small scale in which we extracted for each question the list of URLs that can be the origin of an appropriate text.

We evaluated manually, the quality of our corpus of pairs of questions and texts by calculating the number of URLs retrieved correctly (list of URLS that contain the answer) compared to the total number of URLs used for each question. Our tool achieved a precision of 0.60. In order to improve the quality of our corpus, we carried out a step of filtering which eliminates all words in other languages, any kind of transliterations, etc.

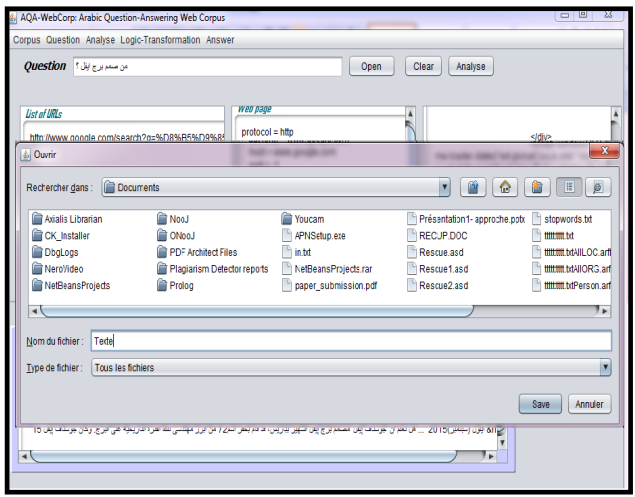

Figure 9.   Saving the text

TABLE I.
STATISTIC OF OUR CORPUS AQA-WEBCORP

| Number of questions | Total of URLs | Number Urls/question | Total texts | Total of corrects URLs /question |
|---|---|---|---|---|
| 115 | 2875 | 23→26 | 115 | 13→16 |

## VII. CONCLUSION AND PERSPECTIVES

It is undeniable that the Arabic corpus is certainly important for various applications in automatic natural language processing. This paper presents our first steps in this field towards the construction of a new corpus dedicated to the Arabic question-answering. It also presents our first version of the AQA-WebCorp (Arabic Question-Answering Web Corpus). It incorporates pairs of questions and texts of five categories, including («أخبار ;صحة و طب ;رياضة ;ثقافة ;إكتشافات ;التاريخ والإسلام ;العالم»). The first phase focuses on Google search for documents that can answer every question, while the later phase is to select the appropriate text. In addition, to improve the quality of our current corpus, we propose to solve the problems mentioned above. Our task is still unfinished; we hope that we can continue to advance the construction of our body, so it could be effectively used for various purposes. It remains to perform post processing necessary to prepare the corpus for the second phase of representing these textual data into logical forms that can facilitate the extraction of the correct answer.

The proposed method is effective despite its simplicity. We managed to demonstrate that the Web could be used as a data source to build our corpus. The Web is the largest repository of existing electronic documents. Indeed, as prospects in this work, we have labeling this vast corpus and make it public and usable to improve the automatic processing of Arabic.

A perspective of extending the current work of analyzing the texts retrieved from the web using a logical formalization of text phrases. This would allow segmenting each text in a list of sentences and generating logical predicates for each sentence. In addition, the presentation as predicates should facilitate the use of a semantic database, and allow appreciating the semantic distance between the elements of the question and those of the candidate answer using the RTE technique.

## ACKNOWLEDGEMENTS

I give my sincere thanks for my collaborators Professor Patrice BELLOT (University of Aix Marseille, France) and Mr Mahmoud NEJI (University of Sfax-Tunisia) that i have benefited greatly by working with them.

## AUTHORS

**Wided BAKARI** received her Baccalaureate in 2004 in experimental science from Secondary school SIDI AICH / Gafsa/Tunisia, and her Technician degree in computer science from sciences in 2007 from Higher Institute of Technological Studies of Gafsa (ISETG)/Tunisia. In 2009, she got her second Bachelor degree in computer sciences from Faculty of Sciences of Gafsa/Tunisia, where she also obtained the master degree in Information system and new technologies in 2012 from the Faculty of Economic Sciences and Management of Sfax (FSEGS)/Tunisia. She is currently a member of the Multimedia and InfoRmation Systems and Advanced Computing Laboratory (MIR@CL). She is now preparing her PhD; her research interest includes information retrieval and Arabic Question Answering systems. She has publications in international conferences (Faculty of Economics and Management, 3018, Sfax Tunisia, MIR@CL, Sfax, Tunisia, Wided.bakkari@fsegs.rnu.tn).

**Patrice BELLOT** is a professor at Aix-Marseille University, Head of the project-team DIMAG, Aix-Marseille University - CNRS, UMR 7296 LSIS (F-13397 Marseille, France, Patrice.bellot@gmail.com)

**Mahmoud Neji** is an associate professor at the Faculty of Economics and Management of Sfax, University of Sfax, in Tunisia. He is a member of the Multimedia and InfoRmation Systems and Advanced Computing Laboratory (Faculty of Economics and Management, 3018, Sfax Tunisia, MIR@CL, Mahmoud.neji@gmail.com).